\documentclass[runningheads]{llncs}

\usepackage{mathtools}
\usepackage{multicol}
\usepackage{bbm}
\usepackage{amsmath}
\usepackage{microtype}
\DisableLigatures{encoding=*, family=*}
\usepackage{fixltx2e} 
\usepackage{notoccite}
\usepackage{multirow,color}
\usepackage [table]{xcolor}
\newcolumntype{P}[1]{>{\centering\arraybackslash}p{#1}}
\usepackage{graphicx}
\usepackage[tight,footnotesize]{subfigure}
\usepackage{caption}

\begin{document}

\title{ XAI-P-T: A Brief Review of Explainable Artificial Intelligence from Practice to Theory}

\author{Nazanin Fouladgar\inst{1} \and Kary Fr\"amling\inst{1,2}  
}	
\institute{Department of Computing Science, Ume\r{a} University, Sweden,\\
\email{nazanin@cs.umu.se}\\ 
\and 
Aalto University, School of Science and Technology, Finland,\\
 \email{kary.framling@umu.se}
}

\titlerunning{A Brief Review of Explainable Artificial Intelligence}
\authorrunning{N. Fouladgar and K. Fr\"amling}
\maketitle
\begin{abstract}

In this work, we report the practical and theoretical aspects of Explainable AI (XAI) identified in some fundamental literature. Although there is a vast body of work on representing the XAI backgrounds, most of the corpuses pinpoint a discrete direction of thoughts. Providing insights into literature in practice and theory concurrently is still a gap in this field. This is important as such connection facilitates a learning process for the early stage XAI researchers and give a bright stand for the experienced XAI scholars. Respectively, we first focus on the categories of black-box explanation and give a practical example. Later, we discuss how theoretically explanation has been grounded in the body of multidisciplinary fields. Finally, some directions of future works are presented.

\end{abstract}

\section{Introduction}\label{section.Intro}

Machine learning has been immensely applied in different applications and is progressing over time by the advent of new models~\cite{Fouladgar-LSTM:2020}. Nevertheless, the complex behavior of these models impedes human to understand concisely how specific decisions were made. This limitation requires machines to provide transparency by means of explanation. Therefore, one could assign ``black-box'' to the machine learning models, aiming at making decisions, and in contrast ``white-box" to the explained version of these models under the topic of \textit{Explanation AI (XAI)}.

The corpus of research in~\cite{MILLER:2019} recall explanation as a reasoning and simulating process in response to ``what happened", ``how happened" and ``why happened" questions. The questions indicate that explanation could be interpreted in forms of both causal and non-causal relationships. Although causality has had more popularity among AI researchers to solve the algorithmic decision problems, the non-causal relationship has recently appealed the scholars of human-computer interaction field. In fact, XAI is not yet matured enough and there are lots of open rooms for providing explanation in practice and theory. While addressing these two aspects are crucial, there is still lack of a brief and concurrent understanding of how currently the XAI field stands practically and theoretically.

In this work, we give an understanding of XAI in literatures, consisting of three directions. First, we provide a neat categorization of black-box explanation incorporating connections between studies in Section \ref{section.categories}. Later, we introduce a practical example in Section \ref{section.practice}. Finally, we take a look at explanation in theory from the social science perspective retrieved from Miller~\cite{MILLER:2019} work in Section \ref{section.social}. We also guide readers through some open research directions in Section \ref{section.future}.

\section{Black-Box Explanation Categories}\label{section.categories}

Explaining machine learning models has recently become prevalent in the XAI domain. Most of the approaches fit into the general categorization proposed by Mengnan et al.~\cite{Mengnan:2019}. The authors discuss two major interpretability categories: \textit{Intrinsic} and \textit{Post-hoc}. While in the former, the focus is on constructing self-explanatory models (e.g. decision tree, rule-based models, etc.), in the latter, the effort relies on creating an alternative model to provide an explanation of existing models. These two classes are rather broad, yet in a more detailed granularity, each of them are divided into \textit{global} and \textit{local} explanation view. The main concern in the global explanation stands for understanding the structure and parameters of the model. However, in the local view, the causal relationship between specific input and the predicted outcome is mainly unveiled. The schematic diagram of this categorization has been illustrated in Figure~\ref{fig.category} in a blue dot block.

Different approaches have been proposed by researchers, incorporating explanation in each of the four discussed classes. Providing global interpretations in the intrinsic category, the models are usually enforced to comprise fewer features in terms of \textit{explanation constraint}~\cite{Freitas:2014} or to \textit{approximate the base models} by readily interpretable models~\cite{Vandewiele:2016}. In the local interpretation of intrinsic category, \textit{attention mechanisms} are widely applied on some models such as Recurrent Neural Networks (RNNs)~\cite{Kelvin:2015}. The latter mechanisms provide the contribution of attended part of input in a particular decision. In general, there is a risk of scarifying accuracy in the cost of higher explanation in the intrinsic category.

Moving toward post-hoc explanation to keep accuracy and fidelity, in the global view, \textit{feature importance} has been discussed massively~\cite{Altmann:2010}. This approach implies for statistical contributions of each feature in black-box models. In this sense, permutating features iteratively and later investigating how the accuracy deviates, have proved high efficiency in the interpretation of classical machine learning models. In case of deep learning models, the main concern is to find preferred inputs for neurons of specific layer to maximize their activation~\cite{Simonyan:2014}. This technique is called \textit{activation maximization}, whose performance highly depends on the optimization procedure. Provided the local post-hoc explanation, a \textit{neighborhood} of input has been approximated by utilizing interpretable models (white-box) and establishing explanation on the new model prediction~\cite{Ribeiro:2016}.
Putting attention on specific models of the latter category, \textit{back-propagating the gradients} (variants) of outputs with respect to their inputs has been discussed in the deep leaning models~\cite{Lapuschkin:2015}. This process refers to a top-down strategy, exploring more relevant features in prediction.

Apart from generic categorization of machine learning explanations in~\cite{Mengnan:2019} and the aforementioned solutions in the literatures, the formal formulations of the categories have been articulated in~\cite{Guidotti:2018}. In this study, the global and local post-hoc interpretations are recalled by \textit{model explanation} and \textit{outcome explanation} terminologies respectively. Figure~\ref{fig.category} shows these terminologies in a red dot block with respect to their equivalent categories in~\cite{Mengnan:2019}. The red bidirectional arrows illustrate these equivalencies. According to the work in~\cite{Guidotti:2018}, the two explanations are formulated as following respectively:

\begin{figure}
		\begin{center}
			\includegraphics[scale=0.75]{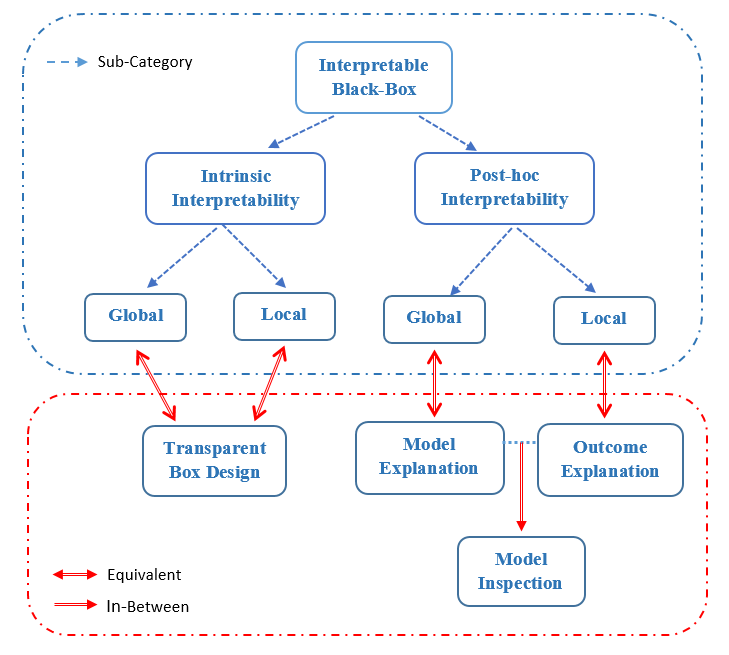}\\
			\caption{Categorization of explainable machine learning models}
			\label{fig.category}
		\end{center}
		\vspace{-6mm}	
\end{figure}

\begin{itemize}
  \item Given a black-box predictor $b$ and a set of instances $X$, the \textit{model explanation} problem consists of finding an explanation $E \in \mathcal{E}$, belonging to a human interpretable domain $\mathcal{E}$, through an interpretable global predictor $c_g = f(b, X)$ derived from the black box $b$ and the instances $X$ using some process $f (., .)$. An explanation $E \in \mathcal{E}$ is obtained through $c_g$, if $E = \varepsilon_g (c_g , X)$ for some explanation logic $\varepsilon_g (., .)$, which reasons over $c_g$  and $X$.
  \item Given a black box predictor $b$ and an instance $x$, the \textit{outcome explanation} problem consists of finding an explanation $e \in \mathcal{E}$, belonging to a human interpretable domain $\mathcal{E}$, through an interpretable local predictor $c_l = f (b, x)$ derived from the black box $b$ and the instance $x$ using some process $f (., .)$. An explanation $e \in \mathcal{E}$ is obtained through $c_l$, if $e = \varepsilon_l (c_l , x)$ for some explanation logic $\varepsilon_l  (., .)$, which reasons over $c_l$ and $x$.
\end{itemize}

Likewise the difference between the global and local post-hoc explanations, the difference between the above formulations relies on the explanations of whole black-box logic and specific input contribution in the black-box decisions. 

Another class of black-box explanation introduced in~\cite{Guidotti:2018} is called \textit{model inspection}, designed to leverage domain-required analysis. This terminology locates somehow in the middle of the two previous categories, the \textit{model explanation} and the \textit{outcome explanation}. In fact, \textit{model inspection} provides a visual or textual representations, to give an explanation of either black box specific property or the decision made (e.g. one could vary the inputs and observe the prediction changes visually in sensitivity analysis). The schematic diagram of this block has been illustrated with a red unidirectional arrow in Figure~\ref{fig.category} and its formal definition is represented as following~\cite{Guidotti:2018}:

\begin{itemize}
  \item Given a black box $b$ and a set of instances $X$, the \textit{model inspection} problem consists of providing a (visual or textual) representation $r = f (b, X)$ of some property of $b$, using some process $f (., .)$.
 \end{itemize}
	
All the three mentioned terminologies in~\cite{Guidotti:2018} stand for defining either an external model or a visual/textual representation. Recalling from~\cite{Mengnan:2019}, the intrinsic explanations are replaced by \textit{transparent box design} terminology in~\cite{Guidotti:2018}. The left bidirectional red arrows in Figure~\ref{fig.category} indicate this replacement and the \textit{transparent box design} is formulated as following~\cite{Guidotti:2018}:

\begin{itemize}
  \item Given a training dataset $D = (X, \hat{Y})$, the \textit{transparent box design} problem consists of learning a locally or globally interpretable predictor $c$ from $D$. For a locally interpretable predictor $c$, there exists a local explanator logic $\varepsilon_l$ to derive an explanation $\varepsilon_l (c, x)$ of the decision $c (x)$ for an instance $x$. For a globally interpretable predictor $c$, there exists a global explanator logic $\varepsilon_g$ to derive an explanation $\varepsilon_g (c, X)$.
 \end{itemize}

As it is clear in the formulation above, the explanation $\varepsilon$ is given based on its own predictor $c (x)$ rather than external global models ($c_g = f (b, X)$) or external local models ($c_l = f (b, x)$).

In addition to the formulations above, the categorized explanation tools such as salient mask (SM) and partial dependence plot (PDP) are scrutinized in~\cite{Guidotti:2018}. This study also investigates the explained machine learning models and their examined data types. It has been turned out that neural networks, tree ensembles and support vector machine are widely explained with tabular data, targeting the \textit{model explanation} category. Yet, the deep neural networks explanation are mostly provided with image data for the purpose of \textit{outcome explanation} and \textit{model inspection}. Solving the \textit{transparent box design} problems, decision rules with tabular data are underlined in the majority of literatures.

\section{Black-Box Outcome Explanation in Practice}\label {section.practice}

As an example of \textit{outcome explanation} in practice, we pinpoint our very recent work in~\cite{Fouladgar:2020}. This paper exploits two neural networks in the classification task of multimodal affect computation over two datasets. Tabular time series data of these datasets are extracted from different wearable sensors such as electrodermal activity (EDA) and learnt by the networks. Accordingly, the networks detect the human state of mind for an individual in each dataset. To explain why a specific state is detected, two concepts of \textit{Contextual Importance (CI)} and \textit{Contextual Utility (CU)}, introduced by Fr\"amling~\cite{Kary:1996}, are employed. 

$CI$ explores how important each sensor (feature) is for a given outcome, i.e. how much the outcome can change depending on the feature value. $CU$ indicates to what extent the feature value of the studied instance contributes to a high output value, i.e. in a classification task how typical the current value is for the class in question. These concepts have been primarily initiated in the context of Multiple Criteria Decision Making (MCDM), where decisions are established on a consensus between different stakeholders preferences. According to~\cite{Kary:2020}, $CI$ and $CU$ are theoretically correct from the Decision Theory point of view. The mathematical formulations of these concepts are as following~\cite{Kary:1996}:

\begin{equation} \label{eq.1}
CI = \frac{Cmax_x(C_i) - Cmin_x(C_i) }{absmax-absmin}
\end{equation}

\begin{equation} \label{eq.2}
CI = \frac{y_{ij} - Cmin_x(C_i)}{Cmax_x(C_i) - Cmin_x(C_i)}
\end{equation}

Where $C_i$ is the $i$th context (specific input of black-box), $y_{ij}$ is the value of $j$th output (class probability) with respect to the context $C_i$, $Cmax_x (C_i )$ and $Cmin_x (C_i )$ are the maximum and minimum values indicating the range of output values observed by varying each attribute $x$ of context $C_i$, $absmax$ and $absmin$ are also the maximum and minimum values indicating the range of $j$th output (the class probability value).

\begin{figure}
		\begin{center}
			\includegraphics[scale=0.55]{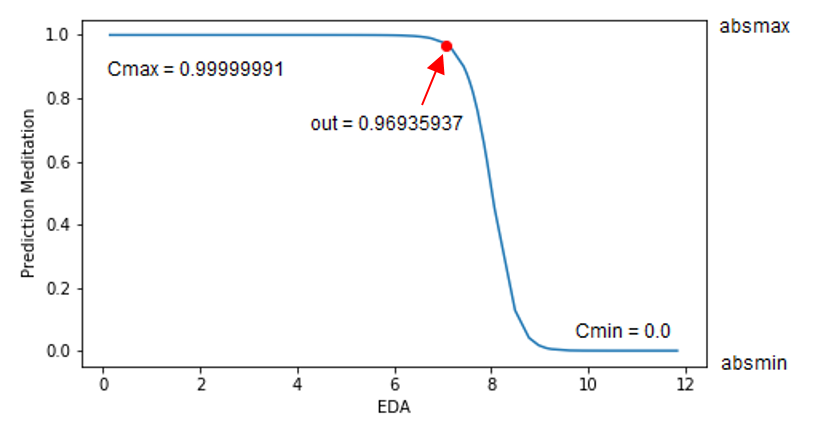}\\
			\caption{$Cmin$ and $Cmax$ values for input variation in EDA~\cite{Fouladgar:2020}}
			\label{fig.practice}
		\end{center}	
		\vspace{-5mm}
\end{figure}

Representing the \textit{outcome explanation}, Figure~\ref{fig.practice} shows the EDA variation for the specific instance ($C_i$) of WESAD~\cite{Schmidt:2018}, one of the examined datasets in~\cite{Fouladgar:2020}. With respect to  this instance, the affective state of an individual has been detected as ``meditation''. As it could be inferred, the $absmin$ and $absmax$ have been ranged in $[0-100]$ and the $Cmin$ and $Cmax$ values are located somewhere within this range, indicating the importance portion of EDA sensor in the process of detecting ``meditation'' state. Considering the EDA sensor's utility, $y_{ij}$ presents a high value within the range of $Cmin$ and $Cmax$ for the context of $C_i$. 


\section{Multidisciplinary Perspectives in Explanation}\label{section.social}

Looking into explanation from a multidisciplinary and theoretical perspective, we highlight Miller's work in~\cite{MILLER:2019}. He scrutinizes explanation as two processes of cognitive and social. While the cognitive process determines a subset of identified causes, the social process mainly aims at transferring knowledge between the explainee and the explainer. Considering explanation as a process and product, the explanation resulting from the cognitive process is appointed to a product. 

In this paper~\cite{MILLER:2019}, one of the key findings from the cognitive science perspective in XAI has been noted as \textit{contrastive explanation}. Such explanation is in congruent with how people explain the causes of events in their daily life. In fact, the main goal is to answer: ``Why $P$ rather than $Q$?”. $P$ refers to a fact did occur and $Q$ is a foil did not occur, yet $P$ and $Q$ follow a similar history. Emphasizing on the why-question, the “within object” differences are mainly under the focus of \textit{contrastive explanation}. However, the question could be designed to cover the ``between objects” or ``within objects over time” differences. In the latter cases, the questions are posed by ``Why does object $a$ have property $P$, while object $b$ has property $Q$?” or ``Why does object $a$ have property $P$ at time $t1$, but property $Q$ at time $t2$?”. 

\begin{figure}
		\begin{center}
			\includegraphics[scale=0.75]{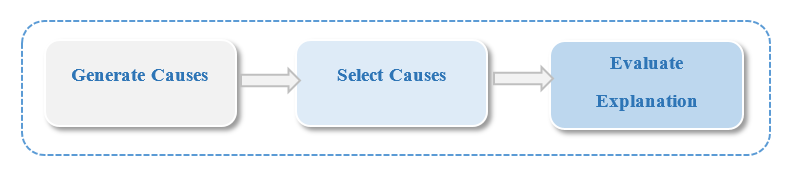}\\
			\caption{Contrastive explanation processes: cognitive science perspective}
			\label{fig.contrastive}
		\end{center}	
		\vspace{-5mm}

\end{figure}

In general, \textit{contrastive explanation} is more applicable than providing the whole chain of causes to the explainee~\cite{MILLER:2019}. Specifying three steps of generating causes, selecting and evaluating, we illustrate the \textit{contrastive explanation} process in Figure~\ref{fig.contrastive}. In the first step, it has been argued that counterfactuals are generated by applying some heuristics. Abnormality, intentionality, time and controllability of events/actions are among these heuristics. In the second step, some criteria are explored to clarify which causes should be selected. Necessity, sufficiency of causes and robustness to changes are a few criteria people typically devote in selecting explanations. Finally, evaluations of provided explanations should be devised concisely. It has been highlighted that people accept more likely the explanations consistent with their prior knowledge. Furthermore, people label explanations good according to the truth of causes. However, the latter measure could not necessarily provide the best explanation. Other measures such as simplicity, relevancy and the goal of explanation could influence the evaluations as well. 

\begin{figure}
\vspace{-2mm}
		\begin{center}
			\includegraphics[scale=0.75]{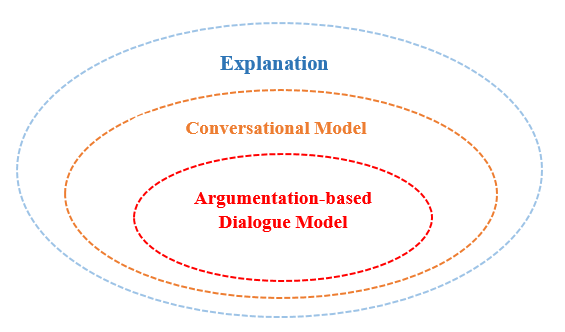}\\
			\caption{Social science perspective of explanation}
			\label{fig.social}
		\end{center}
		\vspace{-6mm}	
\end{figure}

Shifting to the social process, the causal explanations takes the form of \textit{conversation}~\cite{MILLER:2019}. In this process, the communication problem between two roles of ``explainer” and ``explainee” matters and accordingly some rules undertake the protocols of such interaction. Specifically, basic rules follow the so-called \textit{Grice’s maxims} principles~\cite{Grice:1975}: quality, quantity, relation and manner. Clearly speaking, the rules are performed on saying what is believed (quality), as much as necessary (quantity), what is relevant (relation) and in a nice way (manner) to construct a conversational model. In an extension of this model, explanations are presumed  \textit{arguments} in the logical contexts. It has been thought that as well as explaining causes, there should be the potentiality of defending claims. The idea comes from an argument's main components: causes (support) and claims (conclusion). Following the arguments as explanation, a formal explanation dialogue with argumentation framework and the rules of shifting between explainee and explainer have been pointed out in~\cite{MILLER:2019}. One of the advantages of the conversational model of explanation and its extension is their generality, implying for applying in any algorithmic decisions. Being language-independent, these models could be compatible with visual representations of explanation to both explainee questions and explainer answers. To provide how the explanation has been discussed in social perspective, we show the related schematic diagram in Figure~\ref{fig.social}.

\section{Future Works}\label{section.future}

Generally, the literatures discuss concrete categories and formalizations for opening black-box problems, different tools, explained black-boxes and the applied data types. However, there is no explicit categorization of the explained models targeting different types of users such as experts vs. non-experts. Therefore, we believe that a comprehensive review of such models adds values to the XAI research communities. 

In the exemplified work~\cite{Fouladgar:2020}, the $CI$ concept mainly addresses the varying output range of specific variable as a measure of feature importance. There are several cases in multimodal affect computing applications that a particular variable is influenced by other variables of domain. In other words, the current version of $CI$ concept does not consider the casual relationships between endogenous variables. By addressing this gap, one could provide a more realistic and robust black-box explanation to the end-users. Another gap refers to the fact that the $CI$ and $CU$ concepts are not timely dynamic and theoretically well-matured. Therefore, other future directions could be devoted to a timely-aware $CI$ and $CU$ as well as their intersection with the cognitive and social science theories, respectively. 
As Miller~\cite{MILLER:2019} argues the socially interactive explanation, it could also be worthwhile to conduct a user study and investigate the impact of these concepts in a social interaction.


\bibliographystyle{splncs04}
\bibliography{XAI}

\end{document}